\documentclass[letterpaper, 10 pt, conference]{ieeeconf}  
\IEEEoverridecommandlockouts                             
\usepackage{graphics} 
\usepackage{epsfig} 
\usepackage{soul}
\usepackage{xcolor}
\usepackage{hyperref}

\def\BibTeX{{\rm B\kern-.05em{\sc i\kern-.025em b}\kern-.08em
    T\kern-.1667em\lower.7ex\hbox{E}\kern-.125emX}}

\def\BibTeX{{\rm B\kern-.05em{\sc i\kern-.025em b}\kern-.08em
    T\kern-.1667em\lower.7ex\hbox{E}\kern-.125emX}}

\begin{document}

\title{\LARGE \bf
Cancer Subtyping via Embedded Unsupervised Learning on Transcriptomics Data
}
\author{
Ziwei Yang, Lingwei Zhu, Zheng Chen$^{1,*}$, Ming Huang, Naoaki Ono, MD Altaf-Ul-Amin\\and Shigehiko Kanaya
\thanks{This work was supported by the Ministry of Education, Culture, Sports, Science, and Technology of Japan (20K12043) and NAIST Big Data Project and was partially supported by Platform Project for Supporting Drug Discovery and Life Science Research funded by Japan Agency for Medical Research (18am0101111) and Development and the National Bioscience Database Center in Japan.}
\thanks{Ziwei Yang, Lingwei Zhu, Ming Huang, Naoaki Ono, MD Altaf-Ul-Amin, and Shigehiko Kanaya are with Graduate School of Science and Technology, Nara Insitute of Science and Technology, Japan.}%
\thanks{
Zheng Chen is with Graduate School of Engineering Science, Osaka University, Japan.
(e-mail: yang.ziwei.ya3@is.naist.jp; chen.zheng.es@osaka-u.ac.jp)}%
\thanks{
e-mail: yang.ziwei.ya3@is.naist.jp; chen.zheng.es@osaka-u.ac.jp}%
}

\maketitle
\begin{abstract}
Cancer is one of the deadliest diseases worldwide. Accurate diagnosis and classification of cancer subtypes are indispensable for effective clinical treatment.
Promising results on automatic cancer subtyping systems have been published recently with the emergence of various deep learning methods.
However, such automatic systems often overfit the data due to the high dimensionality and scarcity.
In this paper, we propose to investigate automatic subtyping from an unsupervised learning perspective by directly constructing the underlying data distribution itself, hence sufficient data can be generated to alleviate the issue of overfitting.
Specifically, we bypass the strong Gaussianity assumption that typically exists but fails in the unsupervised learning subtyping literature due to small-sized samples by vector quantization.
Our proposed method better captures the latent space features and models the cancer subtype manifestation on a molecular basis, as demonstrated by the extensive experimental results.
\end{abstract}

\section{Introduction}

\subsection{Cancer Subtyping}
Cancer is a leading cause of death worldwide: in 2021 alone it accounted for nearly ten million deaths.
While annually an astronomical amount of budget has been devoted to its treatment and study, there are still several significant obstacles in the way of effective treatment, among which a major one being tumour heterogeneity~\cite{ct1}.
As a result of this heterogeneity, a cancer type usually includes multiple subtypes, while each cancer subtype requires a specific treatment modality.
Therefore, identifying the subtype is a critical step towards a more precise diagnosis and effective treatment for cancer clinically.

Conventional subtype identification is largely based on human experts' knowledge on histopathological and clinical characteristics, such as the tumor morphological appearance, histological grade \cite{ct3}.
Although such heuristic identification has proven their undisputed prognostic value in extensive clinical practice, their limitations are still apparent.
For instance, tumors may have similar histopathological appearance but significantly different clinical manifestation and courses of disease.
In such cases, traditional identification systems rely more on the empirical decision of experts rather than on a uniform clinical basis.
Moreover, due to the lack of standards with a clear molecular basis, traditional identification systems often fall short of more personalized predictions of drug efficacy.

\subsection{Subtyping Alternative with Molecular Information}

Nowadays the booming machine learning (ML), especially deep learning methods have achieved impressive successes in extracting informative patterns or representations from massive data gathered by the virtue of high-throughput technology.
The community has seen several studies featuring the use of ML approaches for automatic cancer type identification and stage diagnosis~\cite{ct4,ct5}.
Experimental validation has proved various genetic mutations are the molecular basis for different cancer subtypes.
The community has shown great interest in identifying the mutations within various kinds of cancers in the hope to accurately subtype cancers to the extent  utilizable for clinical diagnosis.

\begin{figure*}[t]
\centering
\includegraphics[width=0.95\linewidth]{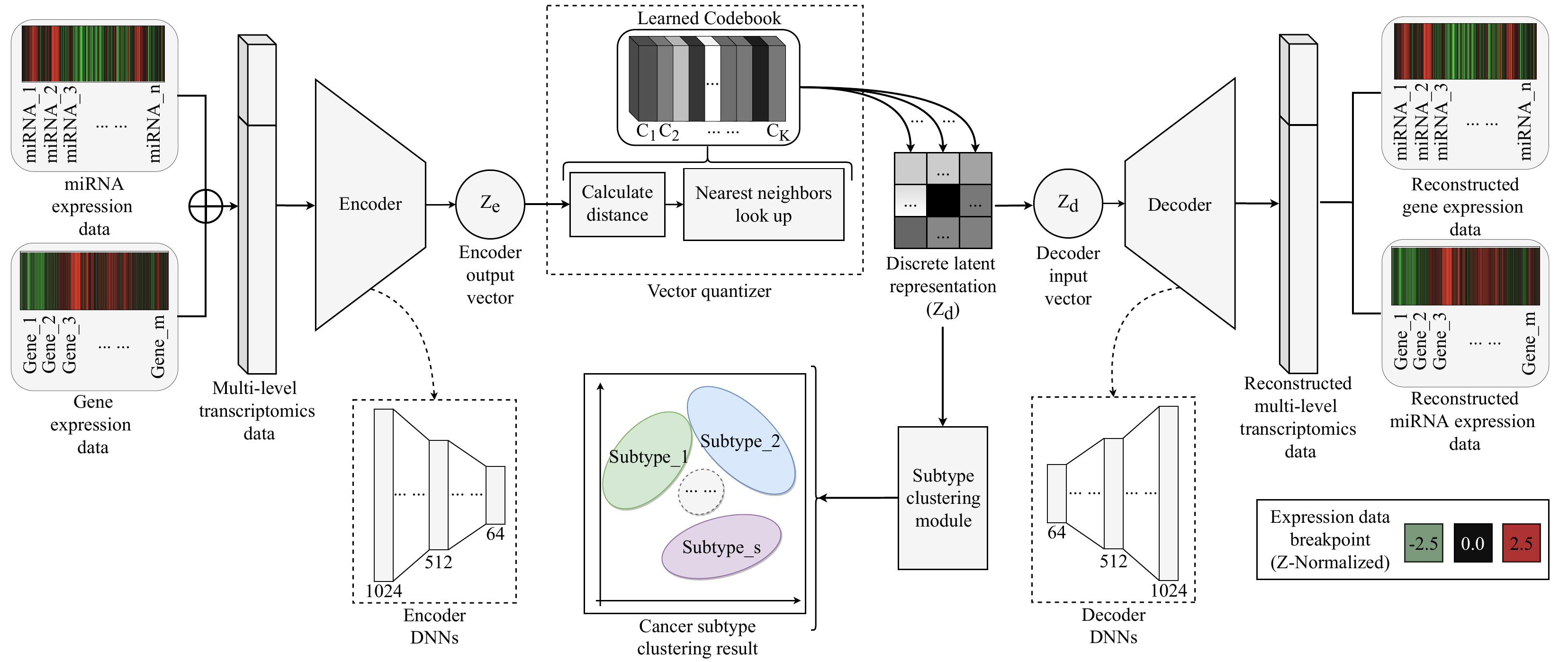}
\caption{The system overview and parameter settings of the proposed model.}
\label{fig:OV}
\end{figure*}

\subsection{Challenges of High-dimensional Molecular Data}

Recent trend of automatic subtype identification is to take as many types as possible of omics data into consideration for more comprehensive/reliable features and subsequently benefiting downstream analysis \cite{ct7}.
However, cancer omics data, especially gene expression data as its basis are normally high dimensional with tens of thousands even hundreds of thousands of molecular features. 
When integrating more omics data like miRNA expression data which may imply the key information about cancer gene regulation and its identification, the input data dimension increases dramatically.
This high dimensionality of data stands as a sharp contrast to the scarcity of data samples: the data used in current studies usually comprises only hundreds of samples from a single cancer type, which implies a severe overfitting phenomenon might exist in trained models~\cite{ct9}.

To overcome the challenges brought by high dimensional but scarce data,
routine methods include applying classic dimensionality reduction algorithms like principal component analysis (PCA) before feeding data into the model. 
However, such forceful reduction may discard some genome-wide hidden patterns when the number of principal axes is small.
Another commonly used practice is performing pre-processing on the original data and retaining only the relevant features, but there is no guarantee such features contain inrrelevant information~\cite{ct10}.
Other researchers proposed to use deep supervised learning methods to automatically  extract relevant information in multi-omics data integration~\cite{ct12}.
These methods often require well-labelled data for learning. 
Unfortunately, in practice especially in cancer subtyping, well-labelled data are precious.

\subsection{Deep Generative Model as a Solution}

The scarce, unlabelled and high-dimensional cancer subtyping data pose a great challenge to supervised learning methods. 
On the other hand, viewing the problem from a different angle might help circumvent the above-mentioned difficulties: instead of only learning the representation from the data itself, it might be more helpful to reconstruct the feature space for subsequent identification.
This observation leads to recent successful unsupervised deep generative models (DGMs) such as the popular variational autoencoder (VAE).
VAE implements variational bayesian inference of the latent data representations and learns the parameters of the underlying distribution. 
Introducing VAE into the cancer omics data integration and feature extracting architectures has been proposed in~\cite{ct16}.
However, VAE fits the data representation in latent spaces to multiple Gaussian distributions, where the Gaussian assumption is for tractability.
Such strong assumption poses a challenge for the model in cancer subtyping tasks, since learning continuous distributions requires ample data.

More recently, Vector Quantised-VAE (VQ-VAE) has been proposed~\cite{ct26}.
VQ-VAE combines VAE with discrete latent representation and gets rid of the imposed assumption of Gaussian prior distribution.
This is implemented by maintaining a discrete codebook which is developed by discretizing the distance between continuous embeddings and the encoded outputs. 
VQ-VAE successfully addresses the shortcoming of VAE and shows superior performance in representing diverse, complex data distributions \cite{ct22}.
To the best of the authors' knowledge, there is no literature discussing the effectiveness of VQ-VAE model on cancer subtyping tasks leveraging cancer omics data yet.

In this study, we propose a VQ-VAE based unsupervised learning model to verify the feasibility of VQ-VAEs to extract a biologically meaningful latent representation of cancer omics data.
The proposed model can integrate multi-level transcriptomics data, namely mRNA and miRNA expression data, and learn discrete latent representations.
We verify the effectiveness of extracted latent representation in identifying cancer subtypes on four cancer datasets.
The results demonstrate that the encoded latent space contains meaningful information of the original cancer transcriptomics data: the latent feature space represents the transcriptome patterns of different cancers.
Hence, it might be safe to put that the VQ-VAE model is promising for analyzing cancer genome data to extract features associated with cancer subtypes.


\begin{table}
\centering
\caption{Cancer dataset used in this study}
\begin{tabular}{cccc} 
\hline\hline
Cancer type & \begin{tabular}[c]{@{}c@{}}Feature size\\(Gene expression)\end{tabular} & \begin{tabular}[c]{@{}c@{}}Feature size \\(miRNA expression)\end{tabular} & Sample size  \\ 
\hline
BRCA        & 13980                                                                   & 319                                                                       & 639          \\
GBM         & 18616                                                                   & 424                                                                       & 417          \\
LGG         & 14262                                                                   & 321                                                                       & 452          \\
OV          & 14231                                                                   & 313                                                                       & 292          \\
\hline\hline
\end{tabular}
\end{table}

\section{Material and Method}
\subsection{Datasets and data processing}
Gene expression and miRNA expression data used in this study were collected from the world's largest cancer gene information database Genomic Data Commons (GDC) portal.
All the used expression data were generated from cancer samples prior to treatment.

Notably, these expression data were contributed from various cancer study projects and institutions, thus normally generated from different assay platforms.
The non-uniform of assay platform implies some technical variations, such as differences in experimental protocols.
This issue causes the downstream expression profile to be easily disturbed by batch effects.
To show the proposed framework is robust against platform diversity and is effective for extensive cancer types, in this study, we used expression data from four cancer types for evaluation:
\begin{itemize}
    \item Breast invasive carcinoma (BRCA): both the  expression data generated from the Illumina Hi-Seq platform and the Illumina GA platform.
    \item Glioblastoma multiforme (GBM): the expression data generated from the Agilent array platform.
    \item Brain lower grade glioma (LGG): the expression data generated from Illumina Hi-Seq platform.
    \item Ovarian serous cystadenocarcinoma (OV): both the expression data generated from the Illumina Hi-Seq platform and the Agilent array platform.
\end{itemize}

Considering discrepancies in gene annotations, some expression features are not always available. 
To further reach the platform independence, cross-platform lost features were removed.

To process expression data generated from the Hi-Seq platform, scaled estimates in the gene-level RSEM (RNA-Seq by Expectation-Maximization) files were firstly converted to TPM (transcripts per million) data and then log-transformed.
For the other expression data generated from Illumina GA and Agilent array platform, non-human expression features were firstly removed, then the missing data imputation was applied by using the IMPUTE R package~\cite{ct25}.
Finally, all expression data were Z-score normalized.
As a result, we get matched expression data from 639, 417, 452, and 292 samples for BRCA, GBM, LGG, and OV, respectively.
The detail of the used dataset after preprocessing is shown in Table I.

Additionally, the cancer subtype information of each sample is collected from the GDC clinical data portal, serving as a benchmark for evaluating the proposed workflow.

\subsection{VQ-VAE}

VAE is an unsupervised generative model for estimating distributions in lower dimensional latent feature spaces ($z$) to represent the data itself.
The random variable $z$ is learned from the given input $X$.
The posterior $p_{\theta}(z|X)$ parametrized by weights $\theta$ is modeled by an encoder network.
Since $p_{\theta}(z|X)$ is an intractable posterior distribution, VAE framework proposes a substitutive distribution $q_{\varphi }(z|X)$ to approximate $p_{\theta}(z|X)$ by  variational inference, while $q_{\varphi }(z|X)$ is usually assumed to be a Gaussian distribution $N(0,1)$.
Meanwhile, a distribution $p_{\theta}(X|z)$ is constructed by the input data and then evaluate the latent feature spaces $z$ and to represent the distribution of $X$ itself via a decoder network.

\begin{equation}
p_{\theta}(X)=p_{\theta}(X|z)p(z)
\end{equation}

The loss function of VAE is minimized following the evidence lower bound, is given by:
\begin{equation}
\mathcal{L}(X;\theta,\varphi)=E_{q_{\varphi}(z|X)}[\log p_{\theta }(X|z)]- KL(q_{\varphi }(z|X)||p(z))
\end{equation}
where the KL-Divergence between approximate distribution and the true posterior can be interpreted as the loss of the inference network (encoder), and the generative network (decoder) need to minimize the reconstruction error.

To escape the strong assumption of a Gaussian prior, the VQ-VAE learns to embed the input data $X$ to the latent space $e\in R^{D}$ and reconstruct the new input $X'$ from it.
The network is an encoder-decoder architecture that is the same as VAE.
Each given $X$ passing through the encoder network is compressed to a feature vector and then is discretized by performing a nearest neighbors lookup in a codebook (a set of latent feature spaces) of the embeddings $C = \{e\}^{K}_{i=1}$, where $K$ is the number of elements in the codebook.
The decoder network reconstructs a $X{'}$ from the quantized embeddings.
The loss function of VQ-VAE is given by:
\begin{equation}
\mathcal{L}  = \mathcal{L}_{recons} + \mathcal{L}_{codebook}+\mathcal{L}_{commit}
\end{equation}
where $\mathcal{L}_{recons} \!=\! \left \| X - X' \right \|^{2}_{2} $ is the reconstruction error between the $X{'}$ and $X$.
$\mathcal{L}_{codebook} \!=\! \left \| S_{g}[E(X)]-e \right \|^{2}_{2}$ is the codebook loss bringing codebook embeddings closer to their corresponding encoder outputs, where $S_{g}$ refers to a stop-gradient operation.
$\mathcal{L}_{commit} \!=\! \beta\left \| S_{g}[e]-E(X) \right \|^{2}_{2} $ prevents the output from fluctuating between different code vectors,  $\beta$ is a hyperparameter that determines its scale.

\subsection{Parameter Settings in the Networks}
The system overview and parameter settings of the proposed model can be seen in Fig. \ref{fig:OV}.
We symmetrically adopt three fully connected layers in both encoder and decoder networks.
In this paper, we use the exponential moving averages (EMA) update in training phase following the \cite{ct26}. 
All experiments of this study were implemented with the Pytorch v1.4 framework and conducted on a server with an NVIDIA GeForce RTX 3090Ti GPU.


\begin{figure}[t]
\centering
\includegraphics[width=0.95\columnwidth]{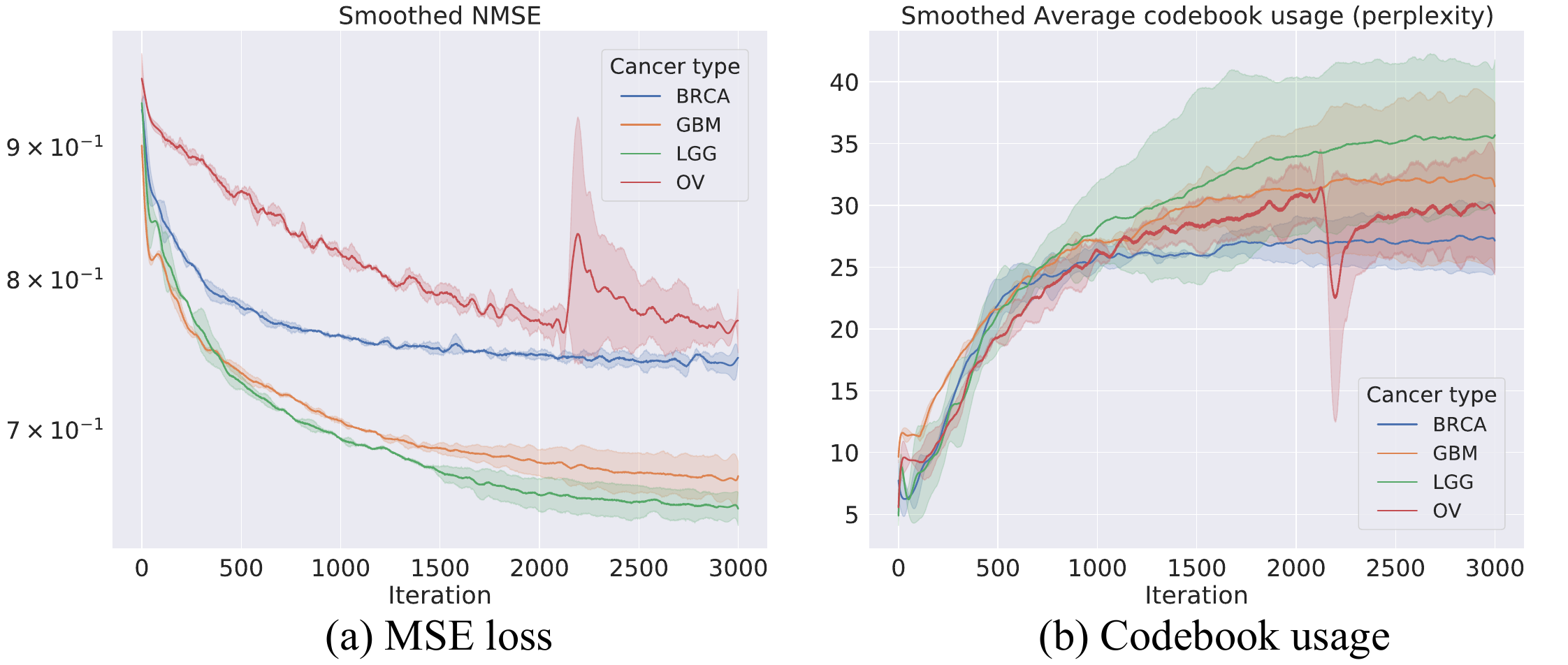}
\caption{Smoothed NMSE and Smoothed Average codebook usage (perplexity)}
\label{fig_mse}
\end{figure}

\begin{figure*}[t]
\centering
\includegraphics[width=0.95\linewidth]{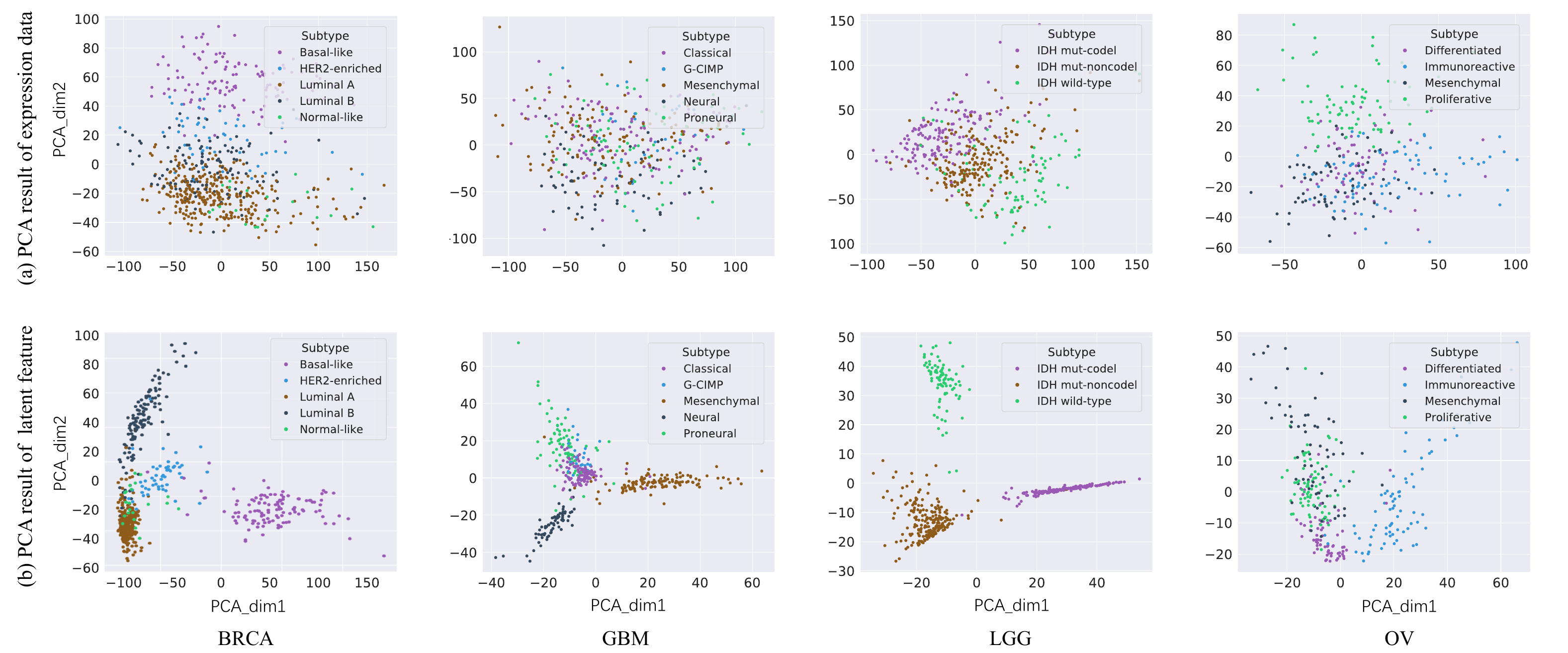}
\caption{PCA visualization of the first two principal axes for four cancer types. 
The upper row shows the clustering result of original expression data and the visualization of latent feature space is illuminated by the lower row.}
\label{fig_pca}
\end{figure*}


\section{Evaluation and Result}
\subsection{Evaluation}
In order to verify the feasibility of this work, for each cancer type, the proposed framework was used to learn the latent representation from the original expression data.
Moreover, to further study whether the learned latent feature can express the biological characteristic of the cancer samples, the principal component analysis (PCA) was performed to reduce the feature dimensionality for more intuitive insight into the latent feature space.
\subsection{Result}
The smoothed MSE loss and average codebook usage during model training are shown in Fig \ref{fig_mse}.
We can see as the number of training iterations increased, the proposed model not only converged with a low reconstruction error but also higher codebook usage.
This demonstrates rich information was passed through the bottleneck layer, thus the model is more likely to learn biologically meaningful representation.

Fig \ref{fig_pca} gives the visualization of samples from four cancer types in the space spanned by the first two principal axes of the original expression data (upper row) and latent feature space (lower row).
We can see samples were much more tightly distributed within their benchmark subtypes in the learned latent feature space than the original expression data space.
Samples from different subtypes also showed distinctive patterns.
More interestingly, we find that the distribution pattern of some cancer subtypes in the latent feature space is identical to their prognosis.
For example, in the breast cancer clinical domain, it is a well-known fact that the Basal-like subtype has the worst prognosis, followed by other subtypes like Luminal A and Luminal B which share most similar molecular characteristics.
These are represented in Fig \ref{fig_pca}: the Basal-like subptype was separated from the other four subtypes by a larger interval.
While in contrast, the boundary between the other subtypes was not clear.
However, these patterns are not significant in the visualization plot of original expression data.

\section{Conclusion}
In this study, we proposed a novel unsupervised learning framework for accurately identifying cancer subtypes.
We observed that the proposed framework successfully learned the biological meaningful latent features from the high dimensional expression data of the cancer samples.
One interesting future direction is to extend the proposed framework to multi-omics data and evaluation of the identified subtype in a further biological context.

\bibliographystyle{IEEEtran}
\bibliography{IEEEabrv,Bibliography}

\end{document}